%% file: main.tex
\documentclass[twocolumn]{svjour3}
\pdfoutput=1
\usepackage{hyperref}

\usepackage{graphicx}
\usepackage{subfig} 
\usepackage{textcomp}
\usepackage{color}
\usepackage{mathtools}
\usepackage{amsmath}

\usepackage{epsfig,graphicx,amsmath,amsfonts}
\usepackage{xcolor,import}
\usepackage{transparent}
\usepackage{upgreek}

\newcommand\numberthis{\addtocounter{equation}{1}\tag{\theequation}}
\usepackage[normalem]{ulem}
\usepackage{multirow}
\usepackage{comment}
\usepackage[]{natbib}

\usepackage{algorithm2e}

\usepackage{tabularx}

\begin{document}
\title{Relaxed-Rigidity Constraints: Kinematic Trajectory Optimization and Collision Avoidance for In-Grasp Manipulation\thanks{Balakumar Sundaralingam was supported in part by National Science Foundation(NSF) Award \#1657596}}
\titlerunning{Relaxed-Rigidity Constraints: In-Grasp Manipulation}

\author{Balakumar Sundaralingam  \and Tucker Hermans}
\authorrunning{Sundaralingam and Hermans}

\institute{University of Utah Robotics Center and the School of Computing,\\ University of Utah, Salt Lake City, UT, USA.\\
\email{bala@cs.utah.edu}}
\date{Received: date / Accepted: date}
\maketitle
\begin{abstract}
This paper proposes a novel approach to performing in-grasp manipulation: the problem of moving an object with reference to the palm from an initial pose to a goal pose without breaking or making contacts. Our method to perform in-grasp manipulation uses kinematic trajectory optimization which requires no knowledge of dynamic properties of the object. We implement our approach on an Allegro robot hand and perform thorough experiments on 10 objects from the YCB dataset. However, the proposed method is general enough to generate motions for most objects the robot can grasp. Experimental result support the feasibillty of its application across a variety of object shapes. We explore the adaptability of our approach to additional task requirements by including collision avoidance and joint space smoothness costs. The grasped object avoids collisions with the environment by the use of a signed distance cost function. We reduce the effects of unmodeled object dynamics by requiring smooth joint trajectories. We additionally compensate for errors encountered during trajectory execution by formulating an object pose feedback controller.
\end{abstract}
\input{introduction}
\input{relatedWork}
\input{problemDefApproach}
\input{extensions}
\input{implementation}
\input{planner_results}
\input{results}
\section{Discussion}
\label{sec:discussion}
We found several open questions to explore from extensive validation of our in-grasp manipulation planner which we discuss below.
\subsection{Improving Manipulation Accuracy}
\label{sec:online-replanning}
Our planner was able to achieve an average position error of 13mm without feedback of the object pose. While this might seem large, there are many tasks that could be performed with this accuracy. One task we explore in this paper is moving a spoon into a cup~(fig.~\ref{fig:c_env}). If only the arm is used to move the spoon, a very precise arm controller or visual servoing is required to move inside the cup. With in-grasp manipulation, the spoon is moved inside the cup without visual servoing, using the dexterity in the fingers. At a broader scale, in-grasp manipulation cannot achieve large object pose changes, as the fingers have limited reachability. We have started exploring methods to switch to a different fingertip grasp to extend the reachable object poses~(\cite{sundaralingam2018regrasp}).

Two potential bottlenecks prevent us from improving the accuracy through online replanning: slow planning time and poor object pose tracking accuracy. Our current trajectory optimization implementation takes on an average 2 seconds to generate a trajectory. The optimization is computationally expensive as the reachability of the fingertips  and the objective function are  highly non-convex. 

This led us to use a Jacobian object pose feedback controller~(Sec.~\ref{sec:feedback}). The feedback controller was unable to reduce the median object position error to less than 1cm. Upon further analysis, we found the object pose tracker was not precise to less than 1cm. We will explore improving the object pose tracking system and study the effect on manipulation accuracy. We will revaluate if the Jacobian controller is sufficient or online replanning is a necessity to improve accuracy.

\subsection{Losing Contact During Manipulation}
Physical experiments showed some of the fingers losing contact on the object during manipulation and making contact again before the manipulation is complete when four fingers were used as seen in Fig.\ref{fig:4f}. This did not lead to dropping of the object. We will be exploring adding tactile feedback to maintain contact with the object. We never observed the object slipping from the grasp during manipulation.

\subsection{Cost vs Constraints}
Our approach formulates the ``relaxed-rigidity'' terms as part of the cost as we want to minimize changes to the initial grasp as much as possible. Another perspective would be to formulate them as inequality constraints with thresholds~(i.e. max allowed deviations). Formulating them as constraints provides a potential advantage of faster planning times. However, finding the thresholds for the ``relaxed-rigidity'' terms that would lead to successful executions on the physical robot is not straightforward. Additionally, a constraint based approach treats all feasible solutions equally while our approach attempts to minimize the deviation when possible.

\section{Conclusion}
\label{sec:conclusion}
We presented an in-grasp manipulation planner, which given only the initial joint angles, the joint limits, and the initial object pose, solves for a joint-level trajectory to move the object to a desired goal pose. We implemented and experimentally validated the proposed method on a physical robot hand with ground truth error analysis. The results show that our relaxed-rigidity constraint allows better real-world performance than assuming a point contact model. We show how to use our planner with a collision avoidance cost to manipulate the grasped object in a cluttered environment. We show the ability to reduce unmodeled dynamic effects by adding a cost for smooth joint space paths. We show that use of an object pose feedback controller reduces the variance in trajectory execution.

\bibliography{rss_2017}
\end{document}

%% file: introduction.tex
\section{Introduction and Motivation}
The problem of robotic in-hand manipulation--changing the relative pose between a robot hand and object, without placing the object down--remains largely unsolved. Research in in-hand manipulation has focused largely on using full knowledge of the mechanical properties of the objects of interest in finding solutions~\citep{Li1989,Mordatch2012,Han1998,Andrews2013}. This reliance on object specific modeling makes in-hand manipulation expensive and sometimes infeasible in real-world scenarios, where robots may lack high-fidelity object models. Learning-based approaches to the problem have also been proposed~\citep{kumar-icra2016,vanhoof-ichr2015-in-hand-rl}; however, these methods require significant experience with the object of interest to work and learn only a single motion primitive~(e.g. movement to a specific goal pose). Solving the general in-hand manipulation problem using real world robotic hands will require a variety of manipulation skills~\citep{bicchi2000hands}. As such, we focus on a subproblem of in-hand manipulation: in-grasp manipulation where the robot moves an object under grasp to a desired pose without changing the initial grasp. We explore a purely kinematic planning approach for in-grasp manipulation motivated by recent successes in kinematic grasp planning~\citep{ciocarlie2007dexterous,carpin2016multi}.

\begin{figure}
  \centering
  \includegraphics[width=0.4\textwidth]{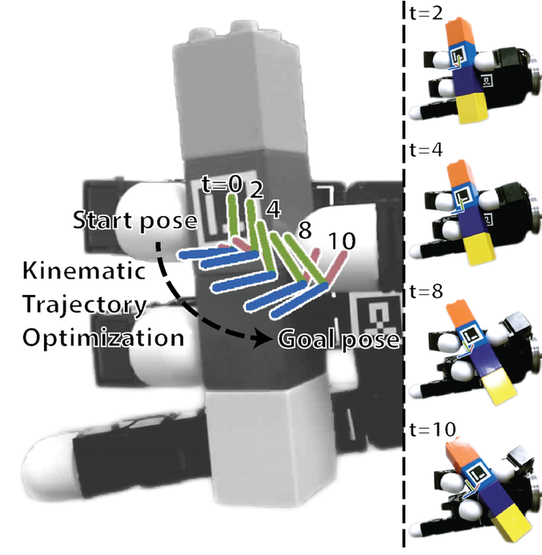}
  \caption{Example trajectory produced by our method executed on the Allegro hand. The trajectory moves the Lego from its initially grasped pose to a desired pose. The robot follows the joint space trajectory produced from our trajectory optimizer with a PD based joint position controller. The images on the right show frames from execution where \(t\) refers to the timestep.}
   \label{fig:intro}
 \end{figure}
Giving robots the ability to perform in-grasp manipulation would allow for changing a grasped object's pose without requiring full arm movement or complex finger gaiting~\citep{Hong1990}. Many tasks requiring a change in relative pose between the hand and the grasped object do not require a large workspace and can be performed without changing the grasp. Example tasks include turning a dial, reorienting objects for insertion, or assembling small parts such as watch gears. This is especially beneficial in cluttered environments where small movements would be preferred to avoid collisions. In this paper, we propose a kinematic planner for in-grasp manipulation through trajectory optimization. The proposed planner gives a joint space trajectory that would move the object to the desired pose without losing grasp of the object. By attempting to maintain the contact points from the initial grasp during manipulation, we do not require any detailed models of the grasped object.

The in-grasp manipulation problem is under-actuated, as the object's states are not fully or directly controllable. As such, it does not immediately offer a kinematic solution. A naive approach would be to model all contacts between the object and robot as rigid links and plan for a desired object pose as if the robot were a parallel mechanism. However, in most robotic hands, the fingers have fewer degrees of freedom (DOF) than necessary to control a 6 DOF world pose. Thus, we introduce a novel cost function which relaxes the rigidity constraints between the object and fingers. This cost function penalizes the robot fingertips for changing the relative positions and orientations between each other from those used in the initial grasp. We name this cost function the relaxed-rigidity constraint. We combine relaxed-rigidity constraints for all fingers with cost terms that encourage the object's  movement to the desired pose. This combined cost function defines the objective for our purely kinematic trajectory optimization. The result allows for small position and orientation changes at the contact locations, while maintaining a stable grasp as the object moves toward the desired pose. Fig.~\ref{fig:intro} shows an example trajectory from our planner. This kinematic planner successfully performed in-grasp manipulation with 10 objects across 500 trials without dropping the object.

Our approach to in-grasp manipulation directly solves for a joint space trajectory to reach a task space goal, in contrast to previous methods~\citep{Mordatch2012,li2013integrating} which rely on separate inverse kinematic (IK) solvers to obtain joint space trajectories. Our direct solution is attractive, as IK solutions become complex when a robot is under-actuated in terms of the dimensions of the task space (i.e. the end-effector of a 4 joint manipulator cannot reach all orientations in a 6 dimensional task space for a given position). Our approach additionally handles hard constraints on the robot's joint positions and velocities. The problem is efficiently solved as a direct optimization using a sequential quadratic programming (SQP) solver. Our method allows for changing the object's pose without the need to know the dynamic properties of the object or the contact forces on the fingers. Solving directly in the joint space also allows us to have costs in the input space such as smooth joint acceleration between time-steps to allow smooth operation of the robot during manipulation. The use of Trajectory optimization also allows for using advancements in collision-free manipulator motion planning~\citep{Schulman2014} to our in-grasp manipulation problem and we show how our planner can avoid collisions with the environment during manipulation. In addition, we compensate for error in trajectory execution online by incorporating an object pose feedback control scheme.

Our ``Relaxed-Rigidity'' planner makes the following contributions validated with real-world experiments:
\begin{enumerate}
\item We demonstrate that a purely kinematic trajectory optimization sufficiently solves a large set of in-grasp manipulation tasks with a real robot hand.
\item We enable this kinematic solution by introducing a novel relaxation of rigid-contact constraints to a soft constraint on rigidity expressed as a cost function. We name this the relaxed-rigidity constraint.
\item Our method directly solves for joint configurations at all time steps, without the need of a separate inverse kinematics solver, a novel contribution over previous trajectory optimization approaches for in-hand manipulation (e.g.~\cite{Mordatch2012}).
\item We are the first to extensively validate an in-grasp manipulation planner on a real robot hand. We do so with multiple objects from the YCB dataset~\citep{Calli2015} and introduce relevant error metrics, paving the way for a unified testing scheme in future works (c.f. Section~\ref{sec:exp}).
\end{enumerate}

This articles makes the following contributions over our previous work~\citep{sundaralingam2017relaxed}:
\begin{enumerate}
\item We introduce a joint acceleration cost to prefer smooth joint space paths, leading to lower object dynamics excitation.
\item We enable collision-free manipulation planning of the object in cluttered environments by including a signed distance cost function.
\item We compensate for error online during trajectory execution through an object-pose feedback controller.
\end{enumerate}

We organize the remainder of the paper as follows. We discuss in-hand manipulation research related to our approach in Section~\ref{sec:related_work}. We follow this with a formal definition of the in-grasp manipulation problem and a detailed explanation of our in-grasp planner in Section~\ref{sec:prob_def}. We present our extensions over the initial planner in Section~\ref{sec:extensions}. We then discuss implementation details and define our experimental protocol in Section~\ref{sec:exp}. We analyze the results of extensive robot experiments in Section~\ref{sec:results}. We discuss the limitations of our approach in Section~\ref{sec:discussion} and conclude in Section~\ref{sec:conclusion}.

%% file: relatedWork.tex
\section{Related Work}
\label{sec:related_work}
In-hand manipulation has been studied extensively~\citep{Li1989,bicchi-icra1995,fearing-ijra1986, Hartl1995}. The topic is often referred to as dexterous manipulation (e.g.~\cite{Han1998}) or fine manipulation (e.g.~\cite{Hong1990}). We choose the term in-hand manipulation to highlight the fact that the operations happen with respect to the hand and not the world or other parts of the robot. We believe that dexterity can be leveraged for a number of tasks, which do not fundamentally deal with in-hand manipulation, and that a robot can finely manipulate objects without the need for multi-fingered hands or grasping. This section covers those methods that are most relevant to our approach and does not discuss in detail methods for finger gaiting (e.g.~\cite{Hong1990,rus-icra1992}) or dynamic in-hand manipulation~\citep{srinavasa-iros2005,Bai2014}.

\cite{salisbury1982articulated} explore grasping of objects with different hand designs. \cite{salisbury1983kinematic} explore gripping forces on grasped objects with three finger, three joint hand designs. Their work on force control with tendon driven articulated hands showed the need for dexterity near the end-effector for manipulation of grasped objects.

\cite{Li1989} developed a computed torque controller for coordinated movement of multiple fingers on a robot hand. The controller takes as input a desired object motion and contact forces and outputs the set of finger torques necessary to create this change. The controller  requires models of the object dynamics (mass and inertia matrix) in order to compute the necessary control commands. The authors demonstrate in simulation the ability for the controller to have a planar object follow a desired trajectory, when grasped between two fingers. \cite{Hartl1995} factors forces on objects and force to joint torque conversions to perform in-hand manipulation accounting for slippage and rolling. An analytical treatment of dynamic object manipulation is explored with ways for reducing the computations required. 

Han et al.~\citep{han-icra1997,Han1998} attempt in-hand manipulation with rolling contacts and finger gaiting requiring knowledge of the object surface. They solve for Cartesian space finger-tip and object poses. Results for rolling contacts are demonstrated using flat fingertips to manipulate a spherical ball. The robot tracks the end-effector velocities using the manipulator Jacobian to determine the joint velocities.

\cite{bicchi-icra1995} analyze the kinematics of rolling an object grasped between fingers. The authors present a planner for rolling a sphere between two large plates acting as fingers. This is achieved through creating a state feedback law of vector flow fields. All these early methods require extensive details about the object which is hard to obtain in the real world and is inefficient when attempting to manipulate novel objects.

In-hand manipulation research has diverged in terms of approaches. \cite{Mordatch2012} formalize in-hand manipulation as an optimization problem. They solve for a task space trajectory and obtain joint space trajectories for the robot using an IK solver independent of their optimization. The trajectory optimization approach factors in force closure, but uses a joint level position controller to perform the manipulation assuming they have a perfect robot dynamics model to convert end-effector forces to positions. Experimental evaluation is shown only in simulation.

Similar to our approach, \cite{Hertkorn2013a} seek to find a trajectory to a desired object pose without changing the grasp. Their approach additionally solves for an initial grasp configuration to perform the desired motion in space. They discretize the problem by creating configuration space graphs for different costs and use a union of these graphs to choose a stable grasp. They perform an exhaustive search through this union of graphs to find the desired trajectory. Their approach does not scale, even to simple 3D problems, with multi-fingered robot hands as stated by the authors and they show no real-world results. Their method is computationally inefficient as the reported time for finding a single feasible trajectory was 60 minutes for a two fingered robot with 2 joints each. In contrast, we efficiently and directly solve for joint positions in the continuous domain using trajectory optimization.

\cite{Andrews2013} take a hierarchical approach to in-hand manipulation by splitting the problem into three phases: approach, actuation, and release. Their method uses an evolutionary algorithm to optimize the individual motions. This requires many forward simulations of the full dynamical system and does not leverage gradient information in the optimization. Another drawback, as stated by the authors, is that their approach cannot be applied to objects with complex geometry.

\cite{Hang2016} explore grasp adaptation to maintain a stable grasp and compensate for slippage or external disturbances with tactile feedback. They use the object's surface geometry to choose contact points for grasping and when performing finger gaiting to maintain a stable grasp. Their method could be used to obtain an initial stable grasp which could then be used in our approach to move the object to a desired pose.

\cite{li2013integrating} use two KUKA arms to emulate in-hand manipulation with tactile and visual feedback to move objects to a desired pose. The use of flat contact surfaces limits the possible trajectories of the object. The use of a 7 joint manipulator as a finger also allows for reaching a larger workspace than common robotic hands which are mostly limited to 4 joint fingers. The evaluation is limited to position experiments and single axis orientation changes.

\cite{scarcia2015local} perform in-grasp manipulation as a coordinated manipulation problem by adding arm motion planning. They assume a point contact with friction model between the object and the fingertips. They enumerate to obtain the reachability for an object pose. They do not perform extensive experiments with objects. 

\cite{Rojas2016} present a method to analyze the kinematic-motion of a hand  with respect to a grasped object. This tool could be used to find feasible goal poses for an object without changing the current grasp similar to~\cite{Hertkorn2013a}. However the authors are motivated by designing dexterous robot hands and do not perform any planning with their technique.

\cite{kumar-icra2014} examine the use of model-predictive control for a number of tasks including in-hand manipulation. They rely on hand synergies and full models of the robot and object dynamics to compute their optimal controllers. However, they recently built on this approach~\citep{kumar-icra2016} and used machine learning to construct dynamics models for the object-hand system. These models could then be used to create a feedback controller to track a specific learned trajectory. They show results on a real robot hand with a high number of states and actuators. However, their method requires retraining to be used if manipulating a new object or moving to a new goal pose. \cite{vanhoof-ichr2015-in-hand-rl} use reinforcement learning to learn a policy for rolling an object in an under-actuated hand. The resulting policy leverages tactile feedback to adapt to different objects, however they must learn a new policy if they were to change the desired goal. Finally, while these learning-based methods show promise in rapidly converging to a desired controller, they still require multiple runs on the robot.

In contrast, we perform in-grasp manipulation on a physical robot hand with novel objects, without requiring extensive object information or performing any iterative learning.


%% file: problemDefApproach.tex
\section{In-Grasp Manipulation Planning through Relaxed-Rigidity Constraints}
\label{sec:prob_def}
\begin{figure*}
  \centering \subfloat[\(t=0\)]{\includegraphics[width=0.38\textwidth]{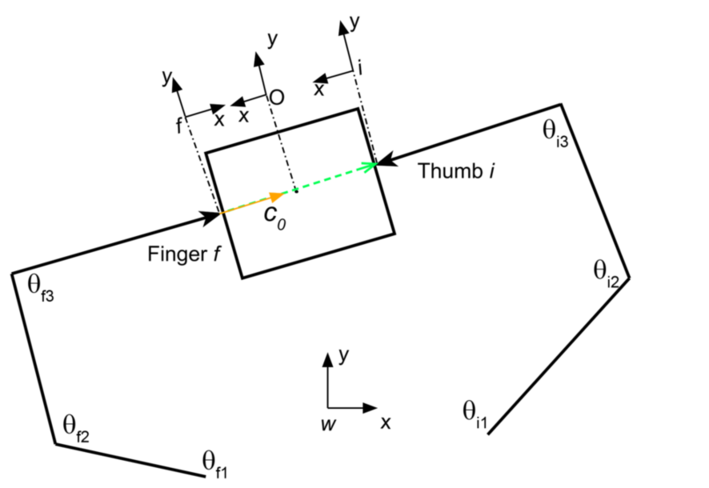}}\hspace{0.1\textwidth}\subfloat[\(t=T\)]{\includegraphics[width=0.38\textwidth]{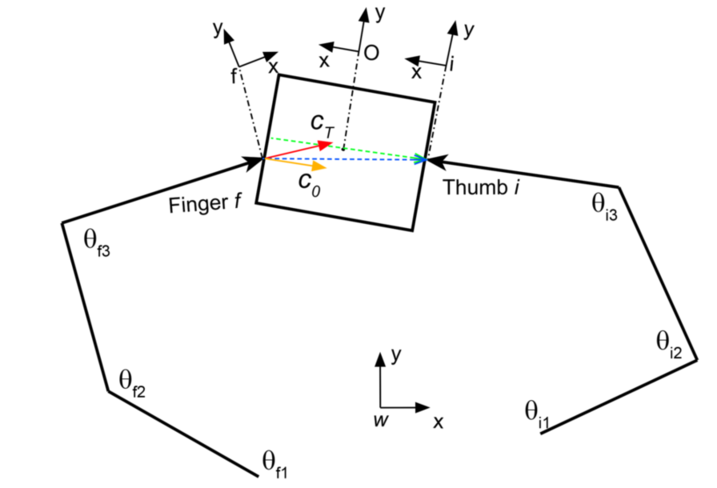}}
  \caption{Depiction of a trajectory optimization solution at the initial and final time steps. The thumb-tip frame is shown as frame \(i\), finger \(f\)'s tip frame is \(f\), and \(w\) defines the world frame. The initial pose of the object with the grasp is shown in (a), the object has to reach the goal pose (b). Our approach models the thumb frame as rigidly attached to the object during the trajectory, while finger \(f\) has a relaxed-rigidity constraint. The effect can be seen in (b), where the relative orientation and position between frames \(i\) and \(f\) have changed from the initial grasp at \(t=0\). The \(\prescript{i}{0}{P}_f\), \(\prescript{i}{T}{P}_f\), \(c_0\) and \(c_T\) terms from Sec.~\ref{sec:relaxed} are shown as green, blue, orange and red vectors respectively. \(\uptheta_{\text{i1-i3}},\uptheta_{\text{f1-f3}}\) are the joint angles of thumb and finger~\(f\).}
  \label{fig:approach}
\end{figure*}
We define the problem of in-grasp manipulation planning as finding a trajectory of joint angles $\mathbf{\Theta}\in [\Theta_1,\Theta_T]$ that moves the object from its initial pose $X_0$ at time $0$ to a desired object pose $X_g$ at time $T$ without changing the fingertip contact points on the object. We address this problem under the following simple assumptions:
\begin{enumerate}
\item The object's pose can only be affected by the robot and gravity, i.e. there are no external systems acting on the object.
\item The object is rigid.
\item The initial grasp is a stable grasp of the object.
\item The desired object pose is in the reachable workspace of the fingertips.
\end{enumerate}
We formulate our solution as a nonlinear, non-convex constrained kinematic trajectory optimization problem:
\begin{flalign*}
\min_{\mathbf{\Theta}}\hspace{5pt}&E_{obj}(\Theta_T, X_g)+k_1\sum\limits_{t=0}^{T-1}E_{obj}(\Theta_t,W_t)\\ &+k_2\sum\limits_{t=0}^{T}E_{pos}(\Theta_t)+k_3\sum\limits_{t=0}^{T}E_{or}(\Theta_t)\numberthis\label{eq:costs}\\
 \text{s.t.}&\\
&\Theta_{min}\preceq \Theta_t \preceq\Theta_{max},\forall t\in[0,T] \numberthis\label{eq:position_limit}\\
&-\dot{\Theta}_{max}\preceq \frac{\Theta_{t-1}-\Theta_t}{\Delta t}\preceq \dot{\Theta}_{max},\forall t\in[1,T] \numberthis\label{eq:velcoity_limit}
\end{flalign*}
The first constraint  enforces the joint limits of the robot hand, while the second inequality constraint limits the velocity of the joints to prevent rapid movements. The scalar weights, $k_1,k_2,k_3$, on each cost term allow us to tune the trade-off between the four cost components. $W_t$ defines the waypoint of the object pose at time $t$ computed automatically as described below.

In order to achieve a purely kinematic formulation, we plan with a number of approximations, which we validate with experiments in Sec.\ref{sec:results}. We now describe the components of the cost function in detail.
\subsection{Object Pose Cost}
\label{sec:obj_pose}
The first term in the cost function~(Eq.\ref{eq:costs}), \(E_{obj}(\Theta_T, X_g)\), is designed to minimize the euclidean distance between the planned object pose at the final time step \(X_T\) to the desired final object pose \(X_g\). Our kinematic trajectory optimization approach assumes no knowledge of the dynamic properties of the object, as such we can not directly simulate the object pose \(X_t\) during our optimization.
Instead, we leverage the fact that in-grasp manipulation assumes no breaking or making of contacts during execution, meaning, in the ideal case, contact points between the robot and object remain fixed.

In our approach we thus plan as if the contact point between the thumb-tip\footnote{The choice of thumb is arbitrary and made only to clarify the discussion. Any fingertip could be chosen to define the reference frame for the object.} and the object is rigid. This allows us to define a reference frame for the object $X$ with respect to the thumb-tip $i$ such that the transformation between the thumb-tip and the object remains fixed during execution. As the thumb-tip  moves with respect to the world frame $w$, we compute the transform to the object frame as
\[
\prescript{w}{}{T}_X =\prescript{w}{}{T}_i\cdot\prescript{i}{}{T}_X \numberthis
\]
where the superscript refers to the reference frame and the subscript to the target frame. The object's transformation matrix is represented by $\prescript{i}{}{T}_X$ with reference to thumb-tip $i$.

We can now transform the desired object pose $X_g$ into a desired thumb pose $G_i$ in the world frame $w$.  The cost function \(E_{obj}\) can now be formally defined as
\[ \prescript{}{}{E}_{obj}(\Theta_T,X_g)=||X_{g}\cdot\prescript{X}{}{T}_{i}-FK(\Theta_T,i)||^2_2 \numberthis\]
where, \(X_{g}\cdot\prescript{X}{}{T}_{i}\) and \(FK(\Theta_T,i)\)
gives the pose of the thumb-tip $i$ with reference to the world frame.  Thus by using the forward kinematics (FK) internally within the cost function we can directly solve for the joint angles of the thumb at the desired object pose.

The second term~\(\sum\limits_{t=0}^{T-1}E_{obj}(\Theta_t,W_t)\) present in the cost function~(Eq.\ref{eq:costs}) encourages shorter paths to the desired pose. We define the waypoints $W_t$ for  every time-step $t$ be linearly interpolated from the initial object pose to the desired object pose $X_g$, equally spaced across all timesteps. We weigh this term at a very low scale relative to the other cost terms, to encourage a shorter path as a linear path is not always guaranteed.
\subsection{Relaxed-Rigidity Constraints}
\label{sec:relaxed}
Since most robotic fingers are under-actuated with respect to possible 6 DOF fingertip poses, we can't apply the same rigid-contact constraint with respect to the object pose, as we did for the thumb for all the remaining fingers. Doing so would reduce the reachable space for the remaining fingers, resulting in a smaller manipulable workspace for the object (manipulable workspace being the workspace covering all possible object poses for a given grasp). Instead, we relax the rigid-contact constraint for all other fingers in the grasp, allowing for a larger manipulable workspace.  The remaining two terms in the cost function~(Eq.\ref{eq:costs}), \(E_{pos}\) and \(E_{or}\), define our novel relaxed-rigidity constraint. The combined effect of the terms encourages the fingertips to remain at the same contact points on the object throughout the trajectory.

We define the cost term $E_{pos}(\Theta_t)$ to maintain the initial relative positions between the thumb, \(i\), and the remaining fingertips, \(f\) throughout execution:
\[
  E_{pos}(\Theta_t)=\sum\limits_{f=1}^{n}||\prescript{i}{0}{P}_{f}-\prescript{i}{}{T}_w\cdot FK_{P}(\Theta_t,f)||_2^2 \numberthis\]
where \(\prescript{i}{}{T}_w\cdot FK_{P}(\Theta_t,f)\) defines the fingertip position for finger $f$ in the thumb frame $i$ at time $t$ and \(FK_{P}(\Theta_t, f)\) computes the position of fingertip \(f\) for joint configuration \(\Theta_t\). Combined with the object pose cost, which moves the thumb towards the goal pose, this cost minimizes deviation from the initial grasp, while moving towards the goal pose.

The last cost term \(E_{or}(\Theta_t)\) encourages the other fingers to maintain their relative orientation to the thumb to be the same as that in the initial grasp. Maintaining this cost across all three orientation dimensions, would again over-constrain the problem to the full rigidity constraint. We relax this constraint by introducing a weight vector $\psi$ which defines a relative preference for deviation in different orientation dimensions
\[E_{or}(\Theta_t)=\sum\limits_{f=1}^{n}||(FK^{i}_{RPY}(\Theta,f)-c^f_0)\cdot\psi||^2_2 \numberthis\]
 where \(FK^{i}_{RPY}(\Theta,f)\) computes the roll, pitch, yaw of the unit vector between the thumb, \(i\) and finger $f$ at time $t$. Fig.~\ref{fig:approach} illustrates the vectors used in the relaxed-rigidity constraints.

%% file: extensions.tex
\section{Extensions}
\label{sec:extensions}
In this section we introduce two extensions to our relaxed-rigidity trajectory planner--joint acceleration smoothness and collision avoidance. We additionally propose an object pose feedback controller to compensate for errors encountered during execution of the in-grasp plan.
\subsection{Joint Acceleration}
We find that the linear interpolation cost term \[\sum\limits_{t=0}^{T-1}E_{obj}(\Theta_t,W_t)\] in Eq.\ref{eq:costs} aids our trajectory optimization in finding a path to the desired object pose; however, it imposes two limitations:
\begin{enumerate}
\item The planner prefers a linear object paths to the desired pose which may not always be possible.
\item The object velocity prefers to be constant during the manipulation which might cause sudden jerk of the object and thereby the joint control.
\end{enumerate}
We explore an alternative cost which prefers smooth paths in the joint space. We minimize the acceleration between time steps following the sum of squares formulation from~\cite{toussaint2017tutorial}. This allows for smoother paths, while also not encouraging the object to follow a linear path to the desired pose.  We replace the linear waypoint interpolation cost term with the following cost term,
\begin{flalign*}
  \alpha_1\sum\limits_{t=0}^{T+1}(||\Theta_{t-2}-2\Theta_{t-1}+\Theta_{t}||_2^2 \numberthis
\end{flalign*}
where we force $\Theta_{T+1}=\Theta_{T}$, and $\Theta_{-1}=\Theta_{-2}=\Theta_{0}$. We discuss the empirical  effects of this in Section~\ref{sec:smoothness-results}.
\subsection{Collision Avoidance}
As proposed in Section~\ref{sec:prob_def} our planner does not avoid collisions between the object and the robot palm or the object and the environment.
We now propose adding an obstacle-based cost function to the optimization in order to obtain collision-free plans while moving the object to the desired pose.
We use signed distance functions to measure the distance between the grasped object and the environment motivated by other trajectory optimization approaches for motion planning~\citep{Zucker2013,Schulman2014}.

The signed distance computes the shortest distance between a point \(p\) and the mesh \(M\). The sign denotes if \(p\) lies within the mesh (negative) or outside the mesh (positive). Given the object mesh, $M$, in the palm frame, the hand joint configuration,~$\Theta$, and the environment as a set of objects,~$W$, the truncated signed distance function can be written as:
\begin{flalign*}
  C(\Theta,M,W)&=\alpha_2\sum_{w\in W}(\beta-\min(\beta,SD(M,w)))\numberthis
 \end{flalign*}
which penalizes the object when it comes within $\beta$ distance of any obstacles in the environment. Ideally collision functions should be used as constrains as in~\cite{Schulman2014}. However, we found that having the collision constraint as a cost term with a large scalar weight~$\alpha_2$ provided better trajectories and quicker solutions. Hence we add this as a cost term. This collision cost can be used to avoid both collisions with the environment and also with the hand. We add this as an additional cost term to Eq.~\ref{eq:costs} and perform trajectory optimization as before.
\subsection{Object Pose Feedback Controller}
\label{sec:feedback}
While our purely kinematic trajectory optimization performs well in practice, it still suffers some error during execution caused by friction, contact dynamics, and other unmodeled effects. Explicitly modeling these variables proves difficult and complex on real-world objects and impossibly to know prior to interaction with a novel object. As such we propose compensating for these errors through feedback controller based on visually tracking the object's pose.
We use as targets the desired object pose trajectory $\mathbf{X}_D$ from initial pose $X_0$ to the desired object pose $X_G$ generated by our ``relaxed-rigidity'' planner.

We define our object pose feedback controller to only affect the thumb joints, as we assume only the thumb attaches rigidly to the object during planning. To ensure the object remains in the robot's grasp, we track the planned joint trajectory $\Theta_D$ for the remaining fingers. As long as the object does not deviate from the planned trajectory by a large margin, thumb-only feedback should prove sufficient to maintain grasp of the object. (We validate this claim in Sec.~\ref{sec:fb_results}.) The robot receives as input the joint position configuration $U[t]$ at every time step $t$. We define this as a combination of the feedforward planned joint trajectory~$\Theta_D[t+1]$ and the object pose feedback term~$\dot{\Theta}_{fb}$,
\begin{align*}
  U[t]&=\Theta_D[t+1]+\lambda_{fb}\dot{\Theta}_{fb}[t] \numberthis
\end{align*}
where the positive weight $\lambda_{fb}$ allows for tuning the feedback compensation.

The feedback input~$\dot{\Theta}_{fb}[t]$ corrects for errors between the planned fingertip pose and the predicted contact pose of the fingertip on the object. The planned fingertip pose at time step $t+1$ is given by $FK(\Theta_D[t+1])$.
The predicted contact pose at time step $t+1$ is computed from the desired object pose~$X_D[t+1]$ and the observed transformation matrix from fingertip to object frame, ~$\prescript{O}{}{\hat{T}}_f$, as
\begin{align*}
  H(X_D[t+1],\prescript{O}{}{\hat{T}}_f)&=Q(R(X_D[t+1])\cdot\prescript{O}{}{\hat{T}}_f) \numberthis
\end{align*}
where $R(\cdot)$ converts a pose into a homogenous transformation matrix and $Q(\cdot)$ transforms a homogenous matrix back into a pose. This essentially accounts for changes in rigid transformation between the object and the fingertip. 

We define our feedback law by transforming the Cartesian space object pose error into the joint space using the inverse of the finger's Jacobian:
\begin{align*}
 \dot{\Theta}_{fb}[t] &= -J^{-1}_{\hat{\Theta}[t]}(FK(\Theta_D[t+1])-H(X_D[t+1],\prescript{O}{}{\hat{T}}_f)) \numberthis
\end{align*}
We found that approximating the Jacobian inverse by its transpose, rather than the Moore-Penrose pseudoinverse performed better for our underactuated fingers.


%% file: implementation.tex
\begin{figure}
  \centering
  \includegraphics[width=0.45\textwidth]{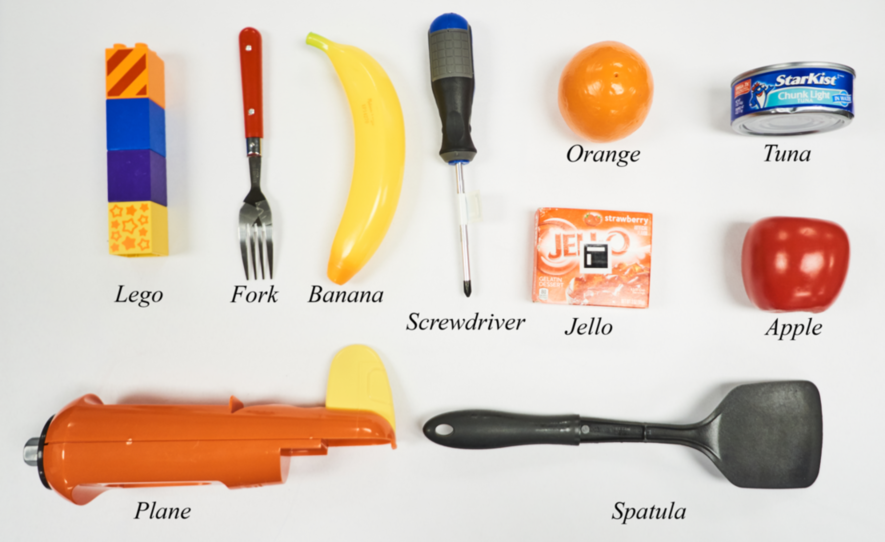}
  \caption{Objects from the YCB dataset with labels below each of them used in our experiments.}
  \label{fig:obj}
\end{figure}
\begin{figure}
  \centering
  \includegraphics[width=0.45\textwidth]{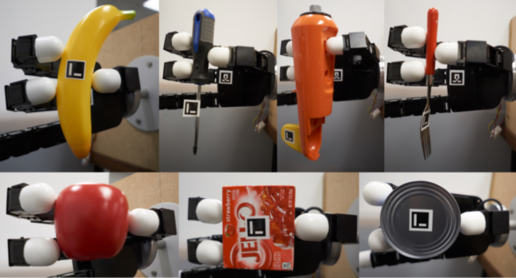}
  \caption{Example grasps tested with various objects in our method. 
  }
  \label{fig:grasps}
\end{figure}
\section{Implementation Details \& Experimental Protocol}
\label{sec:exp}
We now describe important details relating to our implementation and the setup of our experiments.
\subsection{Trajectory Generation and Feedback Implementation}
Direct methods for trajectory optimization, such as sequential quadratic programming (SQP), have shown promising results in robotics~\citep{Schulman2014,posa-ijrr2014,posa-icra2016}. We solve our trajectory optimization problem using SNOPT~\citep{Gill2005} an SQP solver designed for sparsely constrained problems. We run the solver with a maximum limit of 5000 iterations using analytical gradients for the costs and constraints. The computer used to run the solver and experiments is an Intel i7-7700K CPU with 32 GB of RAM running Ubuntu 16.04 with ROS Kinetic~\citep{quigley2009ros}. The robot used is the Allegro Hand, which has four fingers with 4 joints each\footnote{http://www.simlab.co.kr/Allegro-Hand.htm}. We solve for \(T=10\) time steps with each time step being \(\Delta t=0.167\)s long. We expand the obtained solution to a higher resolution of 100 time steps by linearly interpolating the joint trajectories. We limit the joint velocities to be less than 0.6rad/s. Our approach has four weights- three on scaling the importance of each cost term-$k_1,k_2,k_3$ and a projection weight $\psi$ for the orientation cost term $E_{or}$.

The orientation cost $E_{or}$ reduces orientation changes along the weight vector $\psi$. Ideally, we would want to reduce the impact of any contact model on the manipulation task by having weights across all three dimensions. However, this would reduce the reachable workspace of the manipulation task as the Allegro hand is under-actuated with respect to the 6 DOF poses. Hence, we chose to reduce orientation changes along a single axis, which covers the largest workspace of fingertip positions. This is the $y$-axis with respect to the palm for the index, middle and ring fingers, making $\psi=\begin{bmatrix}0& 1 & 0\end{bmatrix}$. We consider this weight as a trade off between allowing a larger manipulation workspace and enforcing smaller changes at contact points. We see in Sec.~\ref{sec:reach-feas-object} that restricting orientation changes along one axis improves the position error over assuming a point contact model.

The remaining three weights model the relative importance between the cost terms of our optimization. We want the robot to always maintain contacts close to  the initial grasp during manipulation, this is taken care of by large value for $k_2$. Keeping the initial orientation is less important, allowing $k_3$ to be less than $k_2$. The weight for waypoints $k_1$ should help guide the fingers to the goal pose, while being low enough to allow for non-linear trajectories when linear trajectories are not feasible. We examined various weights under this scaling and found $k_1=0.09,k_2=100,k_3=1$ to work well across a variety of trajectories and objects. For $k_2<1.0$, the hand dropped the object when unreachable object poses were given. The chosen weights, however, were able to maintain the object in-grasp while still moving the object towards the desired pose. The weights chosen for the extensions are $\alpha_1=0.01,\alpha_2=1000,\beta=0.005,\lambda_{fb}=50$.

For the collision avoidance experiments, the grasped object and the environment are approximately decomposed into convex groups using~\citep{mamou2009simple} to speedup signed distance computation. We compute signed distances using libccd\footnote{https://github.com/danfis/libccd} based on a combination of the Gilbert-Johnson-Keerthi~(GJK) algorithm and the expanding polytope algorithm~(EPA), extensive details are found in~\citep{van2001proximity}. For object pose feedback controller, we use a GPU based particle tracker from~\cite{GarciaCifuentes.RAL} to track the object using a NVIDIA GTX 1060 GPU.
\begin{figure*}
  \centering
  \begin{tabularx}{0.95\textwidth}{>{\setlength\hsize{1\hsize}\centering}X >{\setlength\hsize{1\hsize}\centering}X  >{\setlength\hsize{1\hsize}\centering}X  >{\setlength\hsize{1\hsize}\centering}X  >{\setlength\hsize{1\hsize}\centering}X }
    t=0s & t=0.4s &  t=0.8s &  t=1.2s &  t=1.6s
  \end{tabularx}
  \subfloat{ %
    \includegraphics[width=0.95\textwidth]{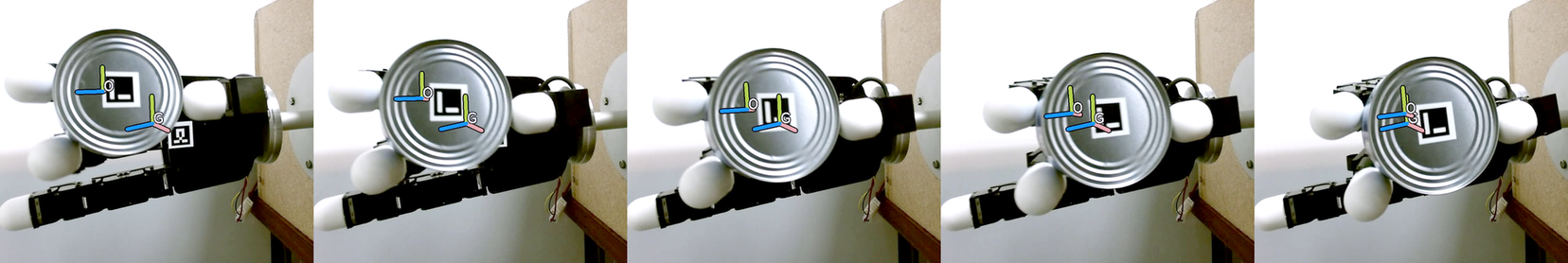}%
  }
  \\[-0.01ex]
  \subfloat{ %
    \includegraphics[width=0.95\textwidth]{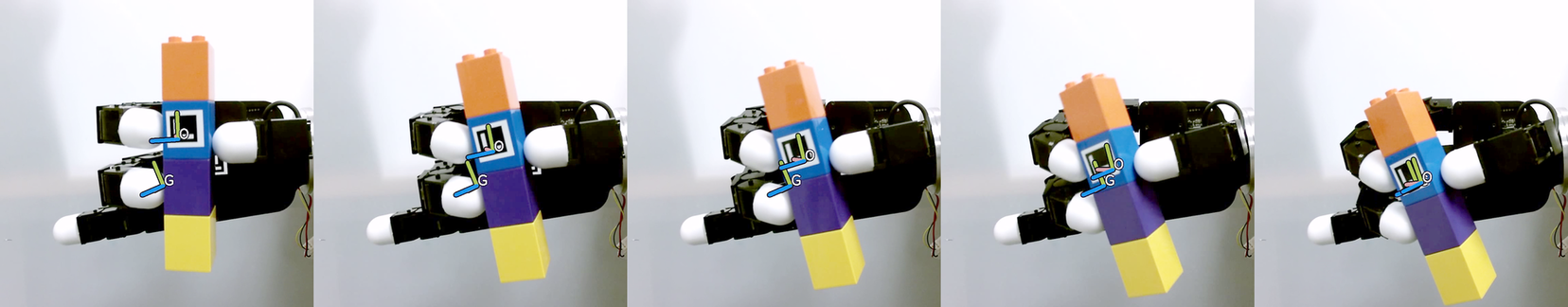}%
  }\\[-0.01ex]
  \subfloat{ %
    \includegraphics[width=0.95\textwidth]{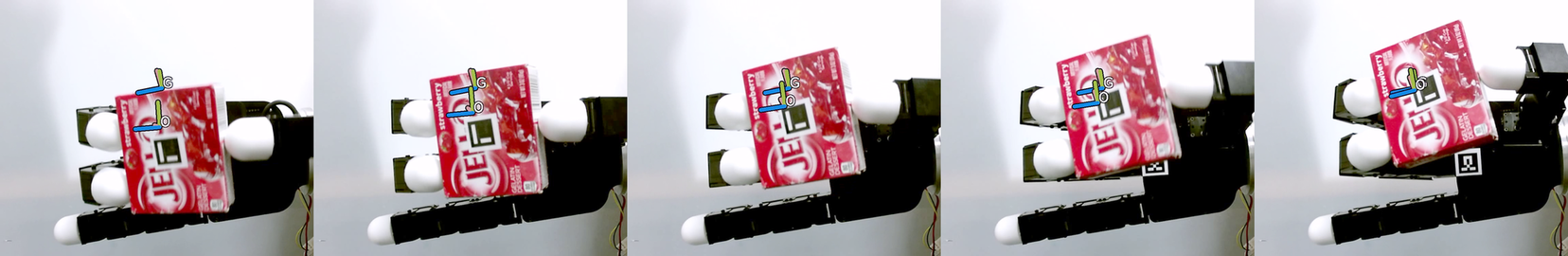}%
  }\\[-0.01ex]
  \subfloat{ %
    \includegraphics[width=0.95\textwidth]{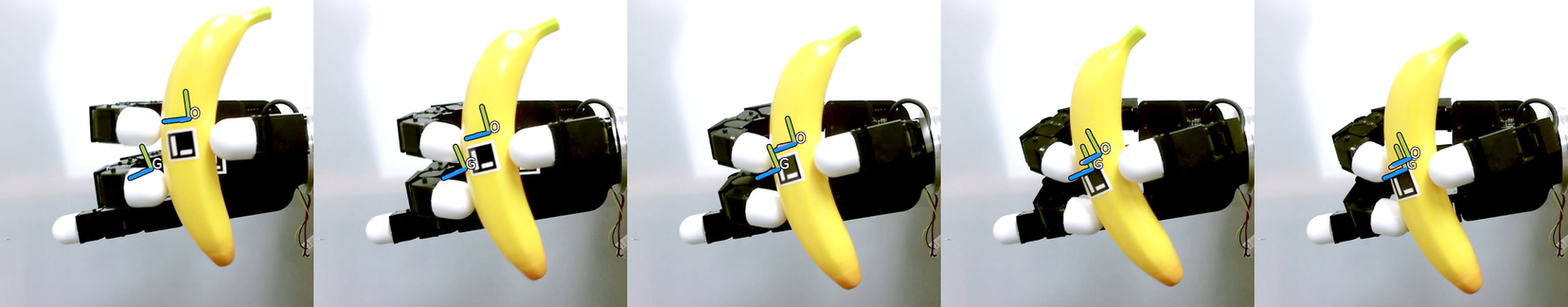}%
  }
  \caption{Images showing manipulation of objects during trajectory execution. The trajectories are generated from our method(``Relaxed-Rigidity''). The frame 'O' represents the current object pose and 'G' frame is the desired object pose. \emph{Tuna} being heavier than all other objects has a larger error due to the PD controller of the hand being insufficient to counteract the gravitational forces. \emph{Banana}, having a complex surface also shows a larger error than objects with a flat surface. Markers used for ground truth collection only. Additional execution of different objects are shown at \url{https://youtu.be/Gn-yMRjbmPE}.}
  \label{fig:manipulation}
\end{figure*}

\subsection{Experimental Protocol}
We selected objects of different size, texture and shape from the YCB dataset~\cite{Calli2015}, shown in Fig.~\ref{fig:obj}, as a benchmarking set.  The ten objects used are: \emph{screwdriver}, \emph{Lego}, \emph{fork}, \emph{banana}, \emph{spatula}, \emph{toy plane}, \emph{Jello}, \emph{tuna}, \emph{apple}, and \emph{orange}. A variety of three-fingered grasps were performed across the objects to show the reachability of the proposed method; examples can be seen in Fig.~\ref{fig:grasps}.

The set of experiments consist of moving the object under grasp to a goal pose. Finding feasible desired poses given an initial grasp is a complex problem~\citep{Rojas2016, Hertkorn2013a} and we do not formalize a method to obtain them. Instead, we focus on obtaining trajectories to a reachable pose and not on finding reachable poses. We obtain goal poses by having a human move the object in-grasp to the desired pose with the robot in gravity compensation mode. Any other method could be used to obtain desired poses. The Euclidean distance to the desired positions from the initial object positions, range from 0.8cm to 8.33cm with a mean of 4.87cm. Desired poses with small positional change have a large orientation change. One trajectory for each goal pose was generated.

The ground truth of the object pose is obtained using Aruco markers~\cite{Aruco2014}. The initial pose of the object is obtained by placing the object in the hand and forming a grasp manually. Once the grasp is set, the joint angles are recorded and the object pose with respect to the palm link is obtained using the Aruco markers. We align the object with the initial pose used for trajectory generation using the markers and robot forward kinematics. Execution of all trials are recorded~(video, robot frames, and object poses). All associated data is available~(\url{https://robot-learning.cs.utah.edu/project/in_hand_manipulation}) to facilitate direct comparison.

Relatively little empirical evaluation has been performed for in-hand manipulation on real robot hands. The lack of a common benchmarking scheme prohibits us from comparing directly with methods described in Sec.~\ref{sec:related_work}. The Allegro hand we use for physical validation has hemispherical fingertips which could cause rolling motion on the grasped object. Modeling rolling motion~(\cite{cutkosky1986friction}) between the fingertips and the grasped object requires extensive information about the object~(surface geometry, friction) and precise force control of the fingertips. The Allegro hand's lack of joint level torque sensing prevents us from comparing our method to methods that use force control. We compare to the ``point contact with friction'' model which can be approximated to a kinematic solution for object manipulation~\citep{Li1989}. We formulate this as a trajectory optimization problem similar to our method with different cost terms. Specifically, we attempt to keep the fingertip positions fixed with respect to the object, while allowing the relative orientation to change. The cost function can be found in~\cite{sundaralingam2017relaxed}. 

We define the following error metrics for evaluating in-grasp manipulation. The position error is computed as the Euclidean distance between the reached position and the desired position of the object. Additionally, we report position error normalized with respect to the length of the trajectory as ``Position Error\%''.

The second metric measures the final orientation error, calculated using quaternions, as the difference between rotation frames is not well defined using Euler angles~\cite{Huynh2009}.
\[err_{orient}=100\times\frac{\min(||q_d-q||_2,||q_d+q||_2)}{\sqrt{2}}  \numberthis\]
where $q_d$ is the unit quaternion of the desired object pose and $q$ is the unit quaternion of the object pose reached. This error is in the range $[0,\sqrt{2}]$ and hence normalized with $\sqrt{2}$ and stated as ``Orientation error\%''. Finally, where appropriate, we report as failed attempts trials where the robot dropped the object during execution.

Ten unique reachable goal poses and two initial grasps per object are chosen to validate our planner.  To account for variation in execution and evaluate robustness, 5 trials are run for each trajectory giving a total of 50 trials per object. The difference in initial position between trials has a mean error of 0.59cm with an associated variance of 0.09cm. A total of 2000 trials are run across different methods to evaluate our proposed method.

To evaluate the joint acceleration extension to our planner and the object pose feedback controller, we conduct experiments with three objects- \emph{Apple}, \emph{Banana} and \emph{Jello}. We choose 5 goals poses per object across two initial grasps per object. We run three trials per generated trajectory. To validate collision avoidance, we show two applications on a physical robot.

 

%% file: planner_results.tex
\section{Results}
\label{sec:results}

We now discuss the results of our empirical experiments.
We first validate our ``Relaxed-Rigidity'' planner on a real robot comparing with alternative formulations for in-grasp manipulation. We then discuss results from our extensions to the ``Relaxed-Rigidity'' planner.
In all plots results correspond to objects grasped with three fingers, unless otherwise stated.
For every trajectory that is run on the robot, the position error and orientation error is recorded. The errors are plotted as a box plot (showing first quartile, median error, third quartile) with whiskers connecting the extreme values. 
\subsection{Relaxed-Rigidity Physical Robot Validation}
\label{sec:reach-feas-object}
\begin{figure}
  \centering
\subfloat{ %
\includegraphics[width=0.48\textwidth]{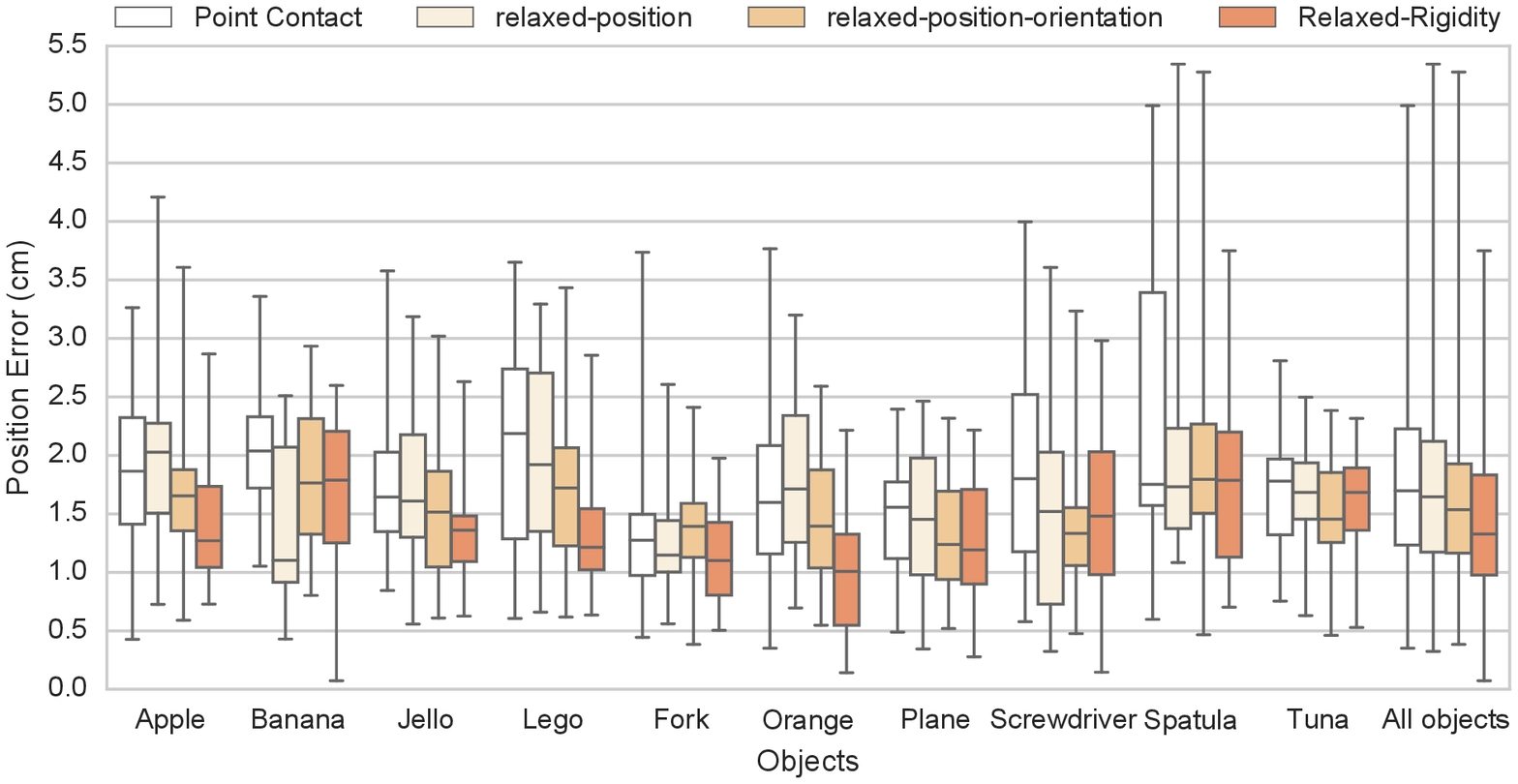} %
}\\[-0.01ex]
\subfloat{ %
\includegraphics[trim={0 0 0 0.73cm},clip,width=0.48\textwidth]{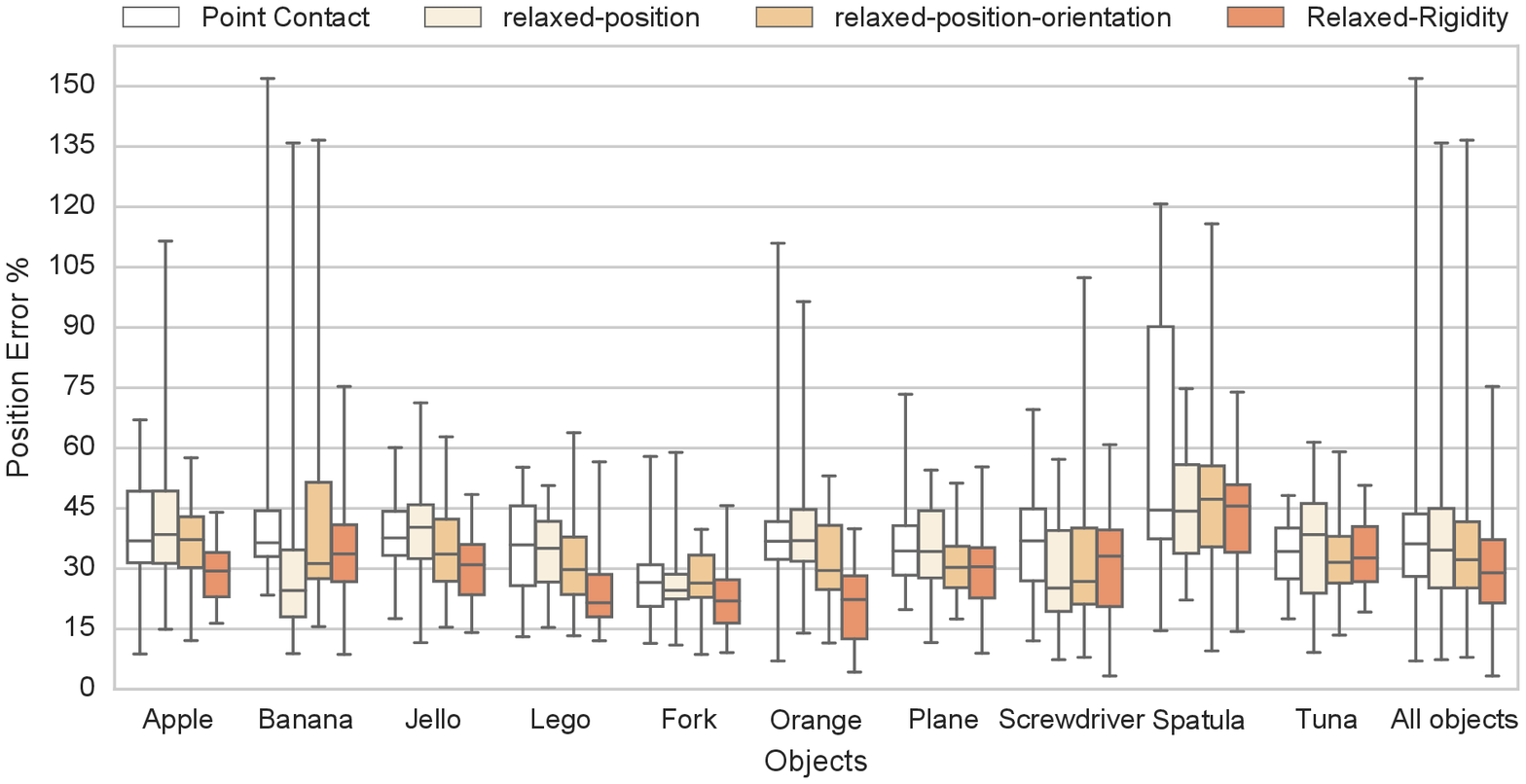} %
}\\[-0.01ex]
\subfloat{ %
  \includegraphics[trim={0 0 0 0.73cm},clip,width=0.48\textwidth]{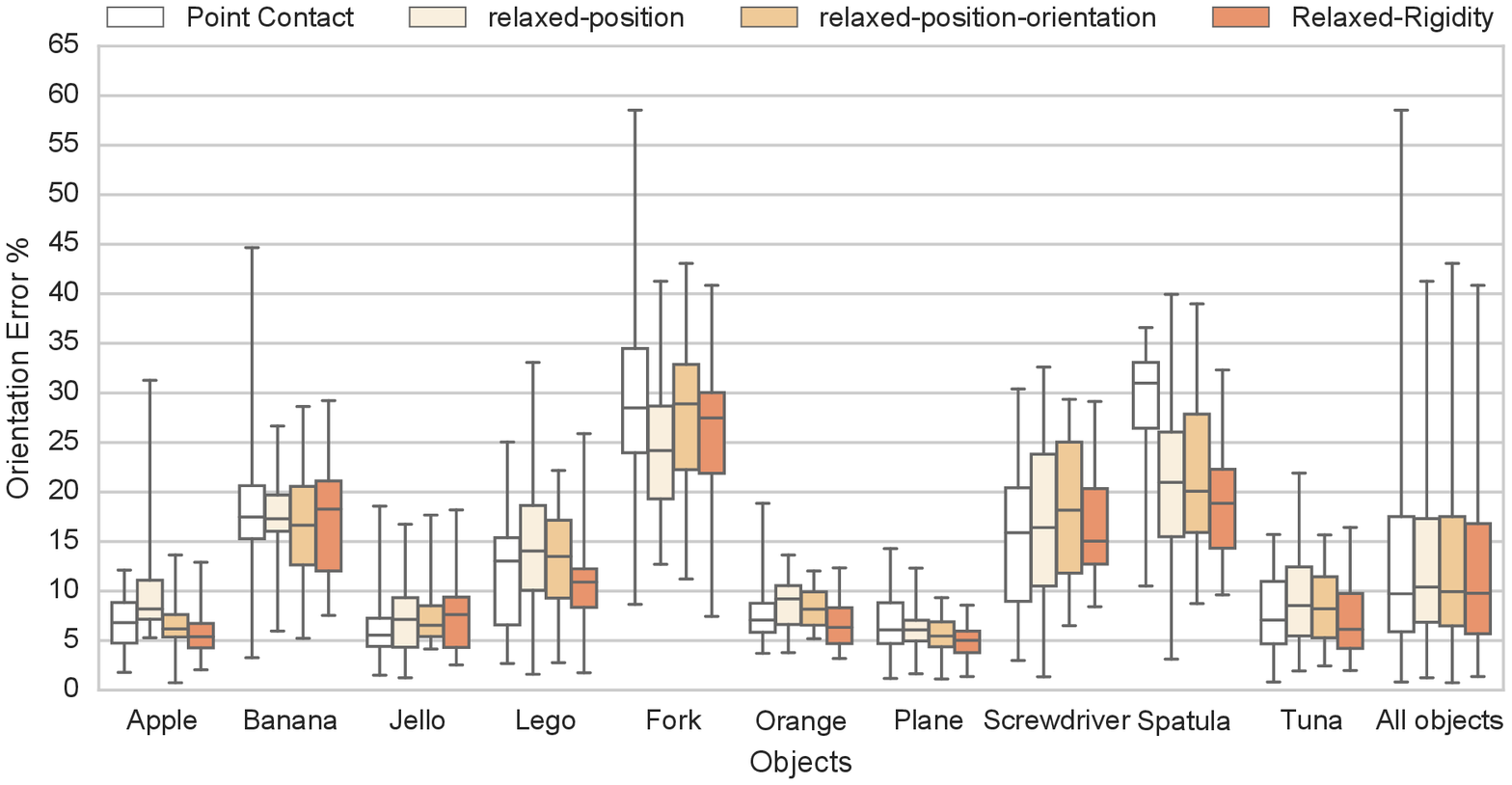}
}
\caption{A comparison of the relaxed-rigidity constraint performance with alternative formulations. Top: Position error Middle: Position error\% Bottom: Orientation error\%. The median position error decreases for all objects with our method. Except for \emph{Banana} and \emph{Jello}, the orientation error\% improves for our method for all objects.}
\label{fig:pose_error}
\end{figure}
The position error and orientation error for all trials across all objects are shown in Fig.~\ref{fig:pose_error}. Our method has the lowest median position error across all objects. The maximum error across all objects is also much smaller for our method than assuming a point contact model with friction. The ``Position Error\%'' plot shows that our method-``Relaxed-Rigidity'' is closer to the desired pose than the initial pose for all trials with a maximum error of 75\%. In contrast ``PC'' obtains error greater than 100\% for several trials, showing the object is moving further away from the desired pose than at the initial pose. Additionally, one can see that our method has a lower variance in final position than the competing methods across nearly all objects. Four samples from our experiments are shown in Fig.~\ref{fig:manipulation} with overlaid current object pose and desired object pose.
\begin{table}
  \centering
  \caption{Summary of results with the best value in bold text. The errors are the median of all trials. ``relaxed-1'' refers to ``relaxed-position'' and ``relaxed-2'' refers to ``relaxed-position-orientation''.}
  \begin{tabular}{|c|c|c|c|c|}
    \hline
\multirow{2}{*}{Method} & \multirow{2}{*}{Suc.\%} & Pos. &\multicolumn{2}{c|}{Error\%}\\ \cline{4-5}
                                 &     &  Error(cm)     & Pos. & Orient.    \\ \hline
    PC                           & 95  & 1.69 & 36.81 & \textbf{9.74}  \\ \hline
    relaxed-1                    & 91  & 1.64 & 30.95 & 10.43 \\ \hline
    relaxed-2                    & 93  & 1.54 & 29.19 & 9.84 \\ \hline
    Relaxed-Rigidity             & \textbf{100} & \textbf{1.32} & \textbf{28.67} & 9.86 \\ \hline
  \end{tabular}
  \label{tab:results}
\end{table}

Table~\ref{tab:results} shows the success rate and the median errors across all these methods. The success rate and position error improve as we add additional costs from our method. It is also seen that our method performs better than assuming a point contact model. The point contact model also resulted in dropping the object on 25 out of 500 trials, while our proposed method never dropped an object.  The orientation error for all methods remains low across all objects except for \emph{Fork} where the fingertips are larger than the object causing it to roll with very small orientation changes at the fingertips. In all objects except \emph{Banana} and \emph{Jello}, the orientation error\%  improves with our method. A large improvement in orientation error is seen in \emph{Spatula}, an object for which the point contact model with friction achieves relatively high orientation error.

\begin{figure}
  \centering
  \includegraphics[width=0.45\textwidth]{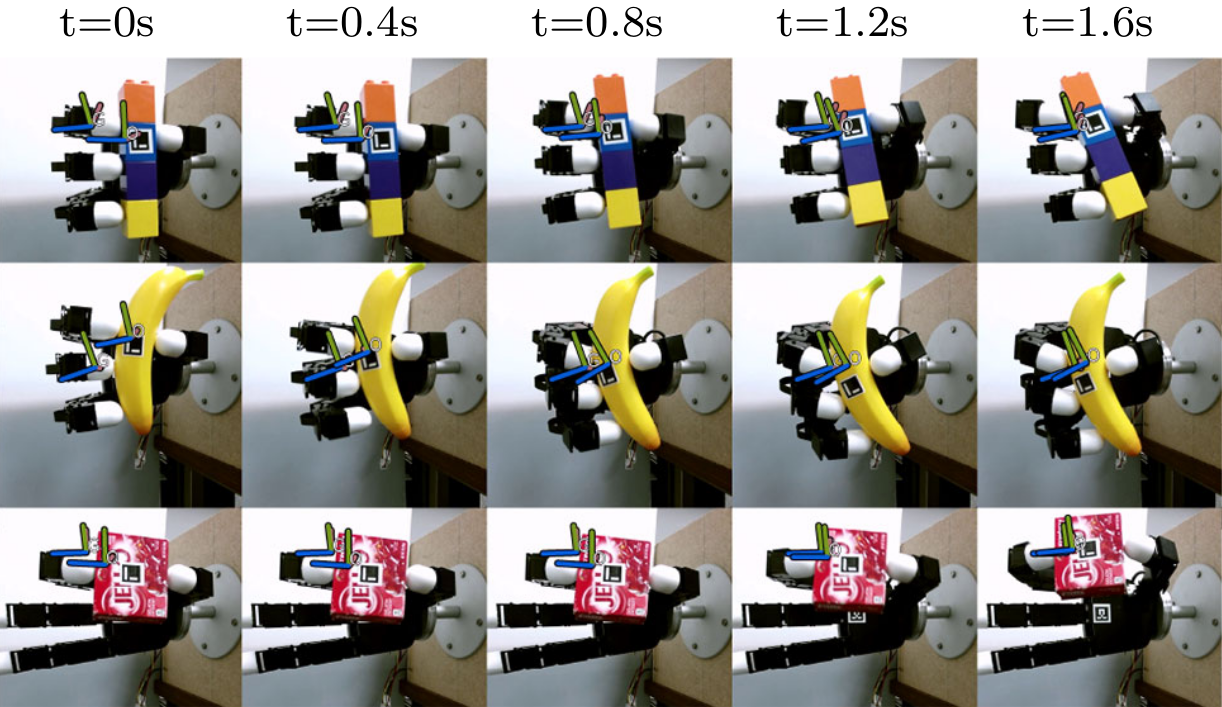}
  \caption{Execution of in-grasp manipulation for four fingered and two fingered grasps. Frame ``O'' is the object pose and frame ``G'' is the goal pose. With the \emph{Banana}, the ring finger loses contact during execution at t=0.4s but makes contact again at t=0.8s and the object reaches the desired pose.}
  \label{fig:4f}
\end{figure}

To show our method generalizes to n-fingered grasps, we show results for 2-fingered and 4-fingered grasps in Fig.~\ref{fig:4f}. We note that 2-fingered grasps tend to shake the object more during trajectory execution than 3-fingered grasps. With 4-fingered grasps, the ring finger sometimes loses and regains contact, adding little benefit over 3-fingered grasps.

%% file: results.tex
\begin{figure}
  \centering
  \includegraphics[width=0.24\textwidth]{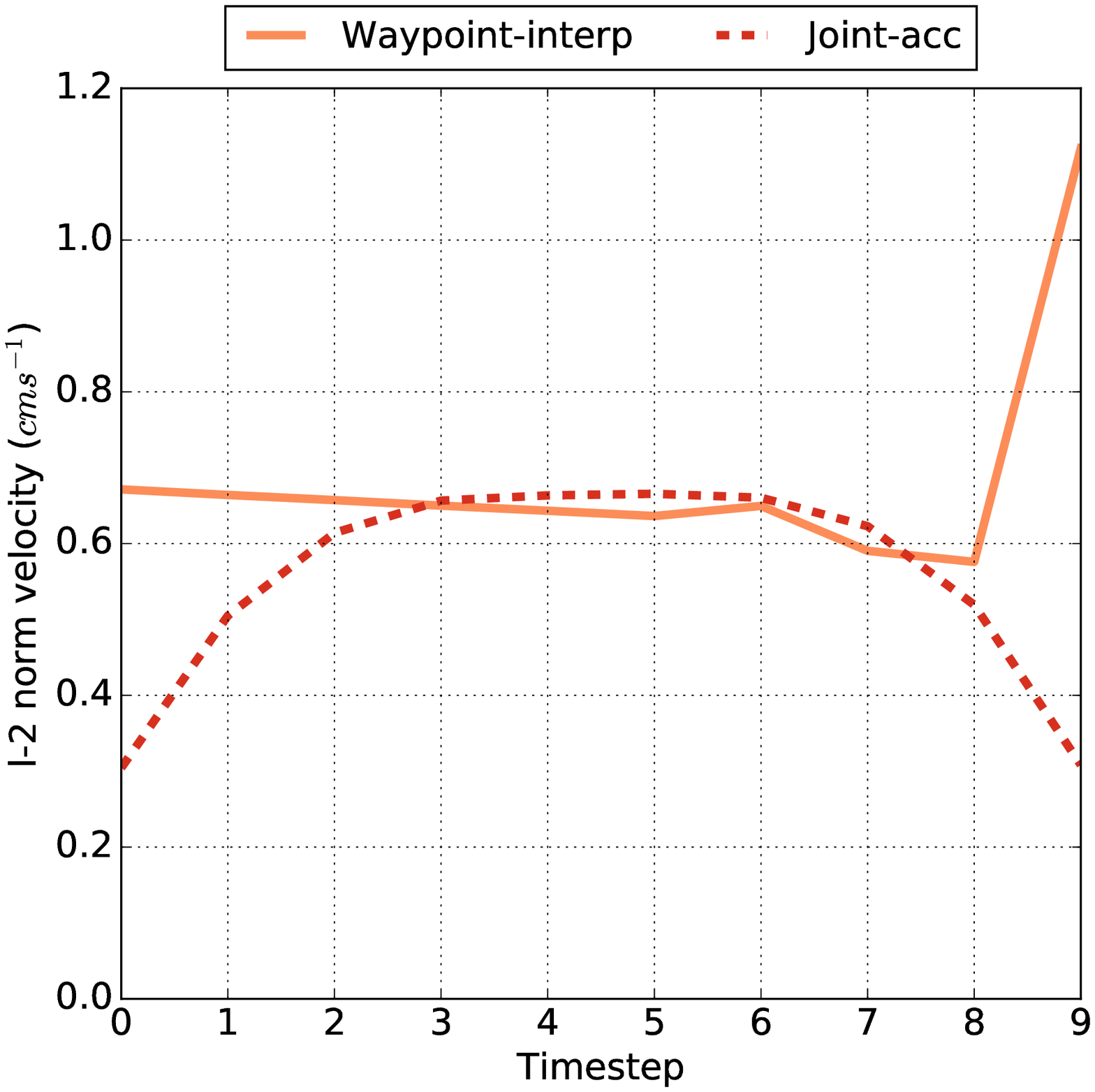}   \includegraphics[width=0.22\textwidth]{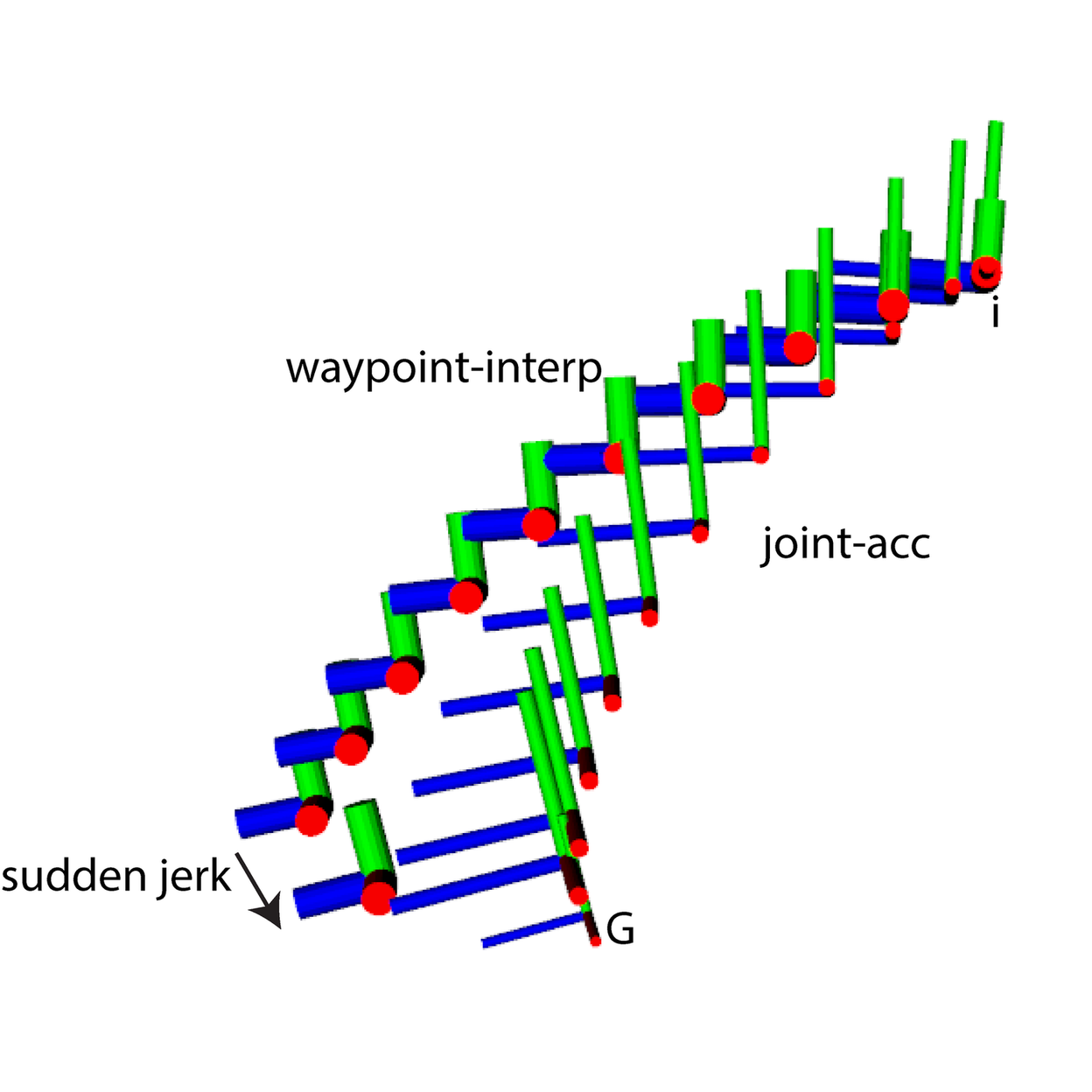}\\
  \begin{tabularx}{0.48\textwidth}{>{\setlength\hsize{1\hsize}\centering}X >{\setlength\hsize{1\hsize}\centering}X }
    (a) & (b)
  \end{tabularx}
  \caption{(a) shows the object velocity from the generated trajectory and (b) shows the object pose trajectory from ``waypoint-interp'' method as fat axes frames and ``joint-acc'' trajectory as slim tall axes frames. The waypoint interpolation method sees a sudden jump at timestep 8 which creates a jerk on the object as seen in~(b). }
  \label{fig:obj_velocity}
\end{figure}
\begin{figure}
  \centering
  \includegraphics[width=0.4\textwidth]{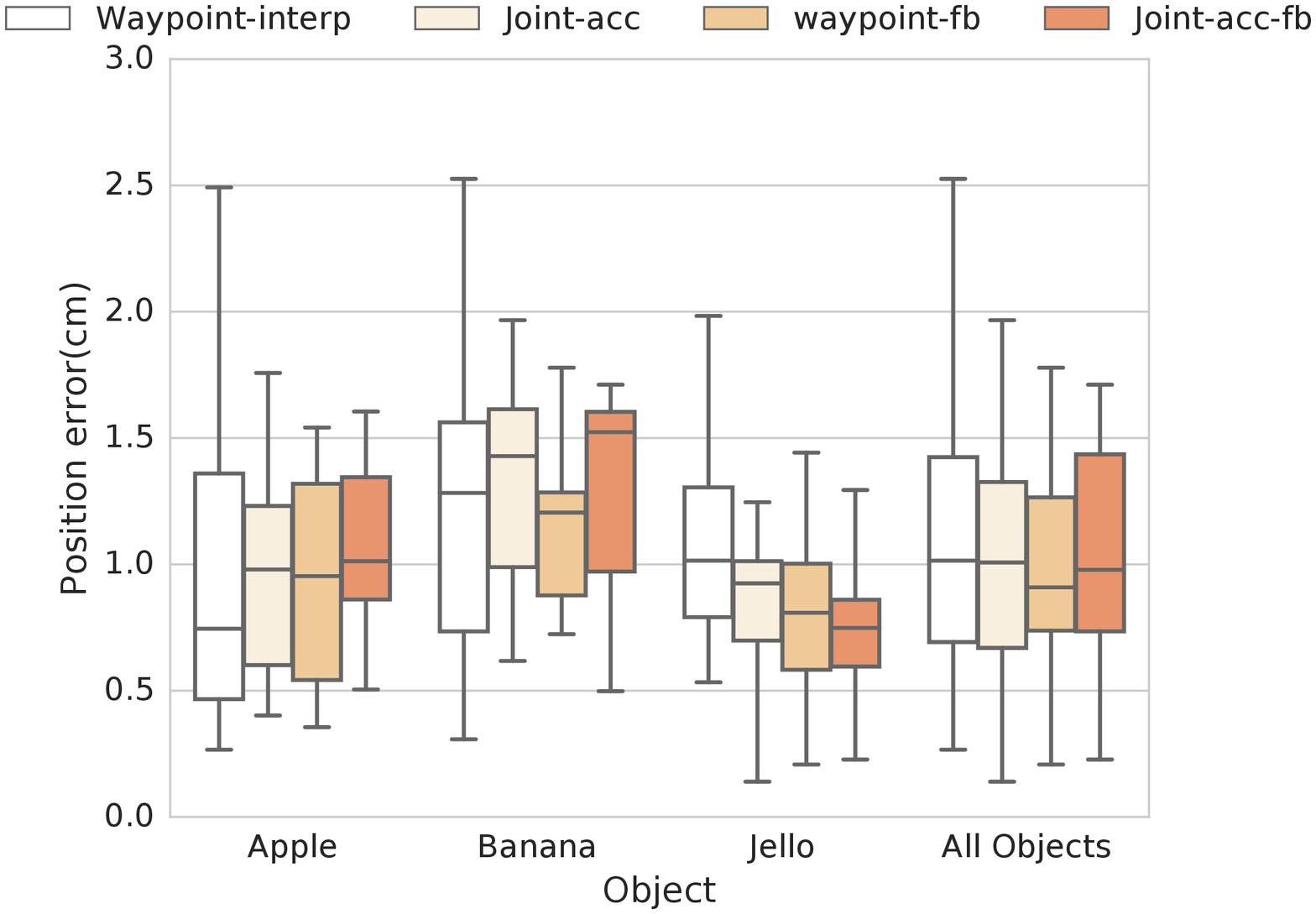}\\
  \includegraphics[trim={0 0 0 1cm},clip,width=0.4\textwidth]{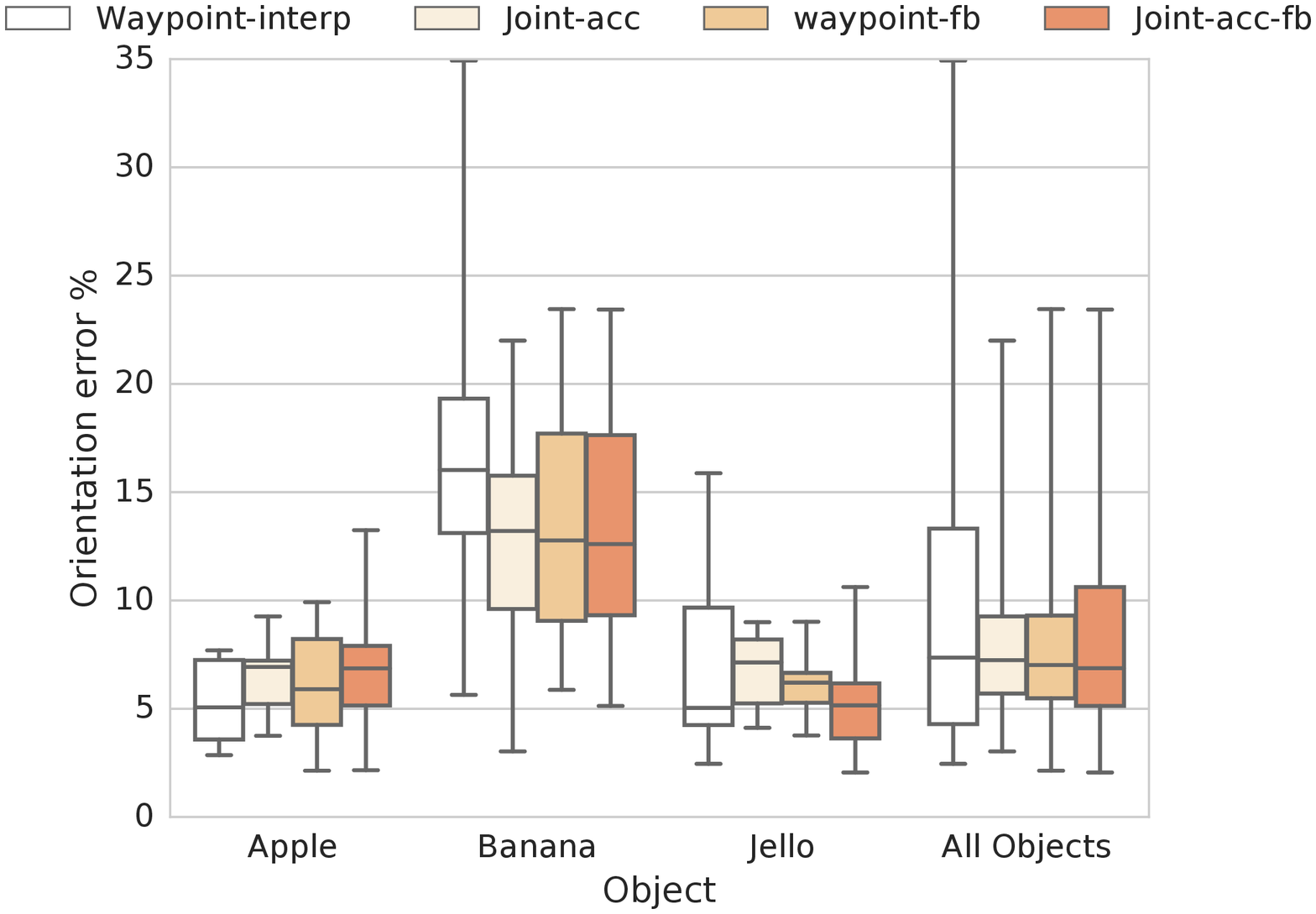}
  \caption{Plots show the position and orientation error in reaching the desired pose across the four methods- ``waypoint-interp'' is our vanilla version, ``joint-acc'' is with our joint acceleration cost, and the methods with suffix ``-fb'' are run with the object pose feedback controller. Variance is reduced with the use of the feedback controller across all objects. Object \emph{Jello} sees a significant reduction in position error due to reduced object dynamics excitation.}
  \label{fig:fb_res}
\end{figure}

\begin{figure*}
  \centering
  \includegraphics[width=0.9\textwidth]{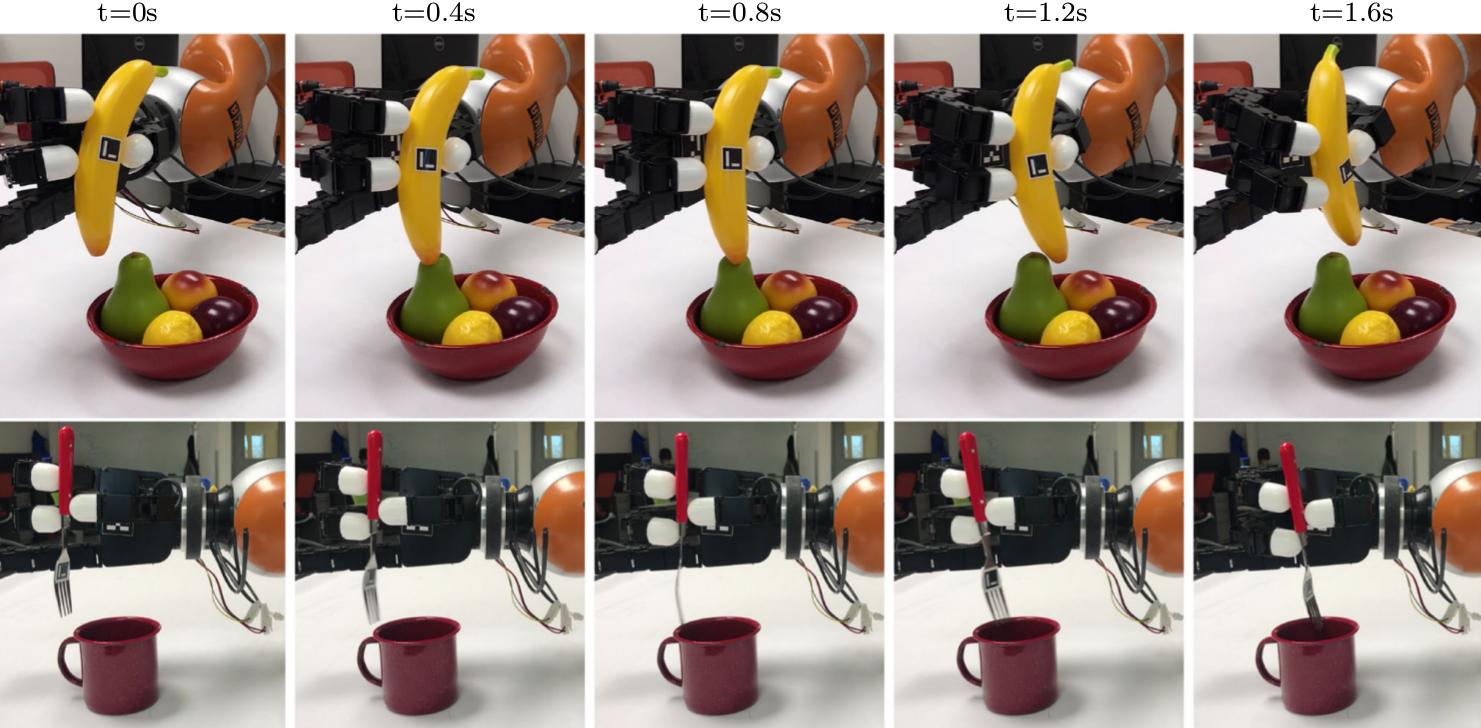}
  \caption{Images showing collision-free manipulation of objects during trajectory execution. The \emph{Banana} object moves around the pear fruit to avoid a collision and reaches its desired pose. The \emph{Fork} object moves to inside the cup avoiding the brim of the cup.}
  \label{fig:c_env}
\end{figure*}
\subsection{Effect of Joint Acceleration}
\label{sec:smoothness-results}
The inclusion of joint acceleration cost term, gives a smooth velocity profile for the object during the in-grasp manipulation as shown in Fig.~\ref{fig:obj_velocity}. Linear interpolation has sudden jerks in the object trajectory if the goal pose is not reachable along the linear path as seen in Fig.~\ref{fig:obj_velocity}. There was no significant difference in planning time and offline convergence errors between the two formulations. However, physical robot validation shows the the joint acceleration generates lower maximum position error and similar median position error to the linear interpolation, as shown in Fig.~\ref{fig:fb_res}. The orientation error for $\emph{Banana}$ sees a significant improvement with ``joint-acc'' as it prevented rolling of the object during manipulation. The \emph{Jello} object sees a significant reduction in position error as the smooth path reduces inertia caused by the powder moving inside the box. We infer the following from our validation: the exclusion of linear interpolation for the waypoint allows for finding a smooth trajectory to the desired pose, the smooth acceleration reduces rolling of the object due to rapid changes to object velocity, and Objects with non-rigidly attached parts have lower error as the smooth acceleration keeps inertia at a minimum.
\subsection{Object Pose Feedback Controller}
\label{sec:fb_results}
We now show results for incorporating the object pose feedback controller on the original relaxed-rigidity planner with linear interpolation costs.
Fig.~\ref{fig:fb_res} shows that the feedback controller drastically reduced the variance in the position and orientation error. We note that nontrivial noise on the object pose persists, caused by the RGB-D based object tracker. This manifests by the lack of error reduction by the feedback controller when error is less than 1cm. Objects with an axis of symmetry such as the \emph{Apple} object proved particularly difficult to track, since the particle filter was unable to find a unique pose.
\subsection{Collision Avoidance}
An interesting application of in-grasp manipulation is to avoid collisions in a cluttered environment by making small changes to the object pose. We setup two such experiments and used our collision avoidance extension to generate trajectories. Fig.~\ref{fig:c_env} shows our in-grasp planner avoiding collisions with the environment while reaching the desired pose. This shows the effectivness of making small changes to the object pose to avoid obstacles in the environment which otherwise would require large motions with the arm. Adding the collision avoidance cost increased the planning time as we compute the signed distance in every iteration between the grasped object and the environment which we decompose into many convex obstacle. It took approximately 120 seconds to generate each collision free plan.
